\documentclass[a4paper, 10pt, conference]{ieeeconf}      

\IEEEoverridecommandlockouts                              

\usepackage{subcaption}
\usepackage{graphicx}
\usepackage{cleveref}
\usepackage{xcolor}
\usepackage{float}
\usepackage{parskip}
\pdfoutput=1 

\newcommand{\KA}[1]{\textcolor{violet}{#1}}

\title{\LARGE \bf 
Learning Decoupled Multi-touch Force Estimation, Localization and Stretch for Soft Capacitive E-skin}

\author{Abu Bakar Dawood$^{1}$, Claudio Coppola$^{2}$ and Kaspar Althoefer$^{1}$, \it{Senior Member, IEEE}
\thanks{This work was supported by a research grant from the Engineering and Physical Sciences Research Council (EPSRC) in the framework of the, National Centre for Nuclear Robotics (NCNR) project (EP/R02572X/1).}
\thanks{This work was completed prior to Claudio Coppola joining Amazon.}
\thanks{$^{1}$School of Engineering and Materials Science, Queen Mary University of London, London E1 4NS, UK
        {\tt\small a.dawood@qmul.ac.uk}}%
\thanks{$^{2}$School of Electronics Engineering and Computer Science, Queen Mary University of London, London E1 4NS, UK.}
\thanks{For the purpose of open access, the author(s) has applied a Creative Commons Attribution (CC BY) license to any Accepted Manuscript version arising.}
}

\begin{document}
\bstctlcite{IEEEexample:BSTcontrol}

\maketitle
\thispagestyle{empty}
\pagestyle{empty}

\begin{abstract}

Distributed sensor arrays capable of detecting multiple spatially distributed stimuli are considered an important element in the realisation of exteroceptive and proprioceptive soft robots. This paper expands upon the previously presented idea of decoupling the measurements of pressure and location of a local indentation from global deformation, using the overall stretch experienced by a soft capacitive e-skin. We employed machine learning methods to decouple and predict these highly coupled deformation stimuli, collecting data from a soft sensor e-skin which was then fed to a machine learning system comprising of linear regressor, gaussian process regressor, SVM and random forest classifier for stretch, force, detection and localisation respectively. We also studied how the localisation and forces are affected when two forces are applied simultaneously. Soft sensor arrays aided by appropriately chosen machine learning techniques can pave the way to e-skins capable of deciphering multi-modal stimuli in soft robots.

\end{abstract}

\section{INTRODUCTION}

Soft robots, unlike traditional, rigid-component robots commonly used in industries, are made from soft, deformable materials and are capable of adapting to the environment in a compliant manner. Because of these properties, soft robots are considered safe when physically interacting with humans  \cite{Luo2017, Lucarotti2015}. Traditional, rigid-component robots have built-in, reliable pose sensors, allowing for the current state of the robot to be easily determined using kinematic models. In the case of soft robots, their compliance and the infinite degrees of freedom they can assume creates challenges in determining their proprioception (sense of their position in 3D space) \cite{Shiva2016}.

In any real environment, a soft robot is likely to physically interact with obstacles. This interaction would lead to the robot deforming, or in other words, experiencing external stimuli. Enabling soft robots to have a sense of their pose (proprioception) and 'awareness' of external stimuli (exteroception) is an emerging research area \cite{Wang2018}. Ideally, embedding the sensors required to enable these capabilities should  not affect the compliance and stiffness of the robot.

Researchers across the world have proposed a range of sensing technologies to determine internal deformations such as stretch and bending in soft robots. Electro-conductive yarn \cite{Wurdemann2015, Wurdemann2015a} and liquid metal filled channels \cite{White2017}, changing the resistance as a function of experienced strain, have been employed for robot shape and stretch sensing. Similarly, capacitive \cite{Larson2016,Glauser2019} and optical sensors \cite{xie2013fiber, searle2013optical} have also been used for sensing strains. Glauser et al. have used capacitive sensing in stretchable skin to measure its deformation. Larson et al. \cite{Larson2016} have used capacitive sensing to examine the effects of stretch, internal pneumatic pressure and external force separately. However, the effect of coupling\KA{, i.e.,} when two or more moduli are applied at the same time, has not been addressed in their work.

\begin{figure}[t]
    \centering
    \includegraphics[width=3.3in]{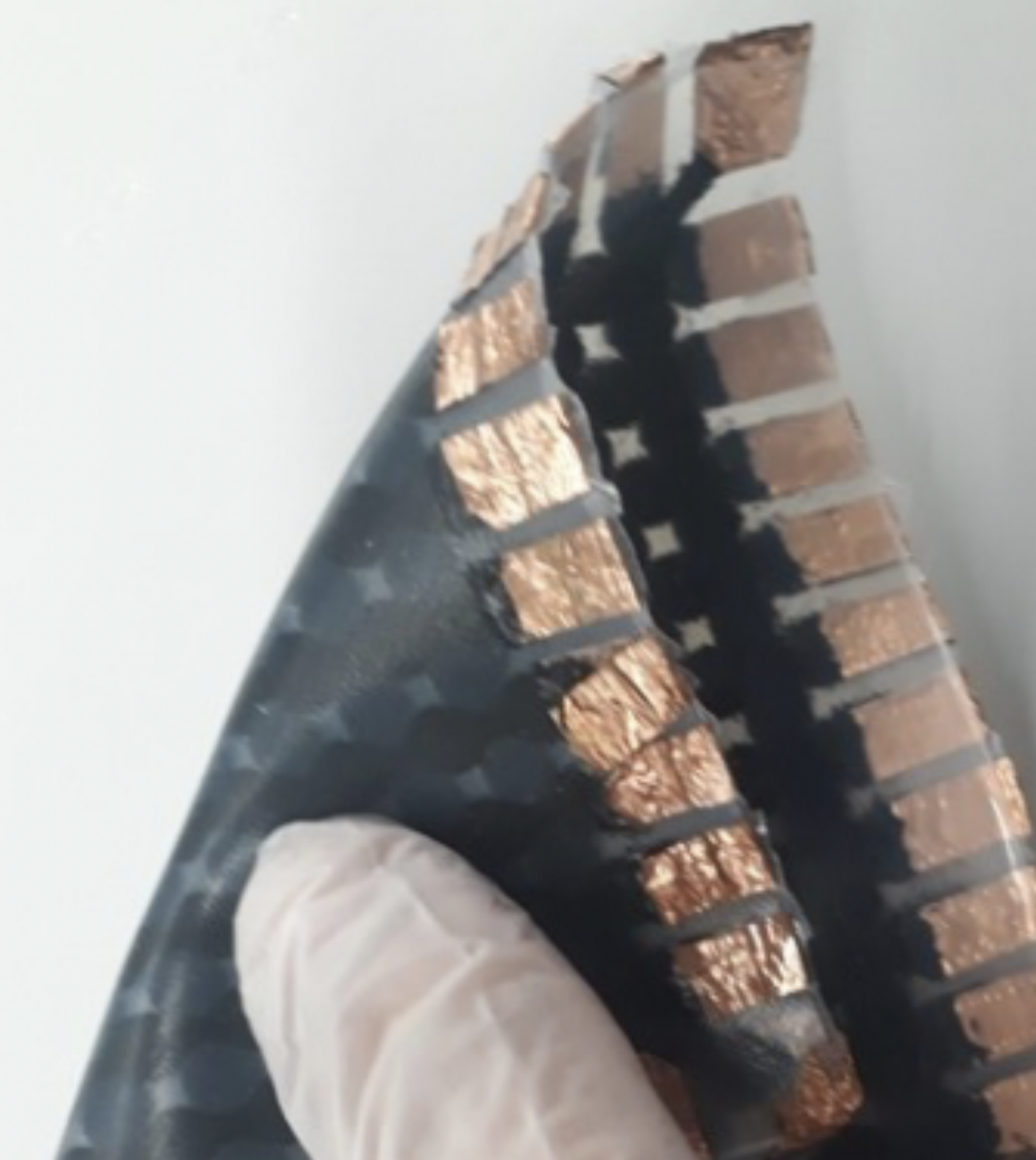}
    \caption{Soft Capacitive e-skin. It is fold-able, highly stretchable and can measure force applied, localization of application of force and stretch.}
    \label{fig:title_image}
\end{figure}

When it comes to multi-modality, combinations of different types of sensors can be employed to detect multiple stimuli. A combination of resistance and photo-sensitivity has been used by Totaro et al. for bending and force estimation \cite{Totaro2017}. Supersonic Cluster Beam Implantation (SCBI) was used to make thin tracks of gold nanoparticles on a silicon substrate and the change in resistance of these tracks was used to determine the extent of bending. The change in the intensity of light transmitted through the substrate was used to determine external forces applied to the sensor. The use of this sensor in soft robots is yet to be explored.

A multi-modal skin was developed by Ho et al. \cite{Ho2016}. This stretchable sensor skin employed thermal, humidity and pressure sensors in alternate layers. Maiolino et al. developed a flexible tactile sensor system using capacitive sensing \cite{Maiolino2012, Maiolino2013}. They integrated a further sensor for temperature compensation of pressure readings. A soft actuator was sensorised by Wall et al. using a combination of liquid metal strain and pressure sensors \cite{Wall2017}. However, measuring multiple stimuli in a soft skin employing a single type of sensor with multi-modal sensing capabilities has not been investigated yet.

In previous work, we have separately studied the changes in capacitance with regard to the application of force, internal pressure and global stretch, as well as coupled changes when multiple stimuli act in unison \cite{Dawood2019}. Extending from this work, we developed a silicone-based capacitive e-skin that uses an array of capacitive sensors. We further developed an algorithm that differentiates between the local and global deformations and hence decouples and estimates stretch, indentation and identifies the location at which the indentation occurs \cite{Dawood2020, Dawood2020a}. However, the heuristic algorithm that we developed uses the maximum change in capacitance to localise indentation, hence becoming ineffective when the skin is subjected to multiple indentations/forces.

To explore scenarios where more than one force is applied to the skin at different locations, we have to find more sophisticated solutions. In this paper, instead of using our heuristic algorithm, we develop a machine learning method to estimate global deformation (i.e. stretch), local deformation (i.e. force) and the localisation of local deformation in situations when up to two forces are simultaneously applied in different locations. The contributions of this paper are:

 
 \begin{enumerate}
     \item A modified experimental setup enabling the skin to measure force instead of indentation.
     \item A data-oriented approach to estimate force, localisation and stretch in case of a single contact experienced by the sensor skin.
     \item Force estimation and localisation when the skin is subjected to two indentations at two different locations simultaneously.

 \end{enumerate}

\section{SOFT CAPACITIVE E-SKIN}

\begin{figure}[t]
    \includegraphics[width=3.35in]{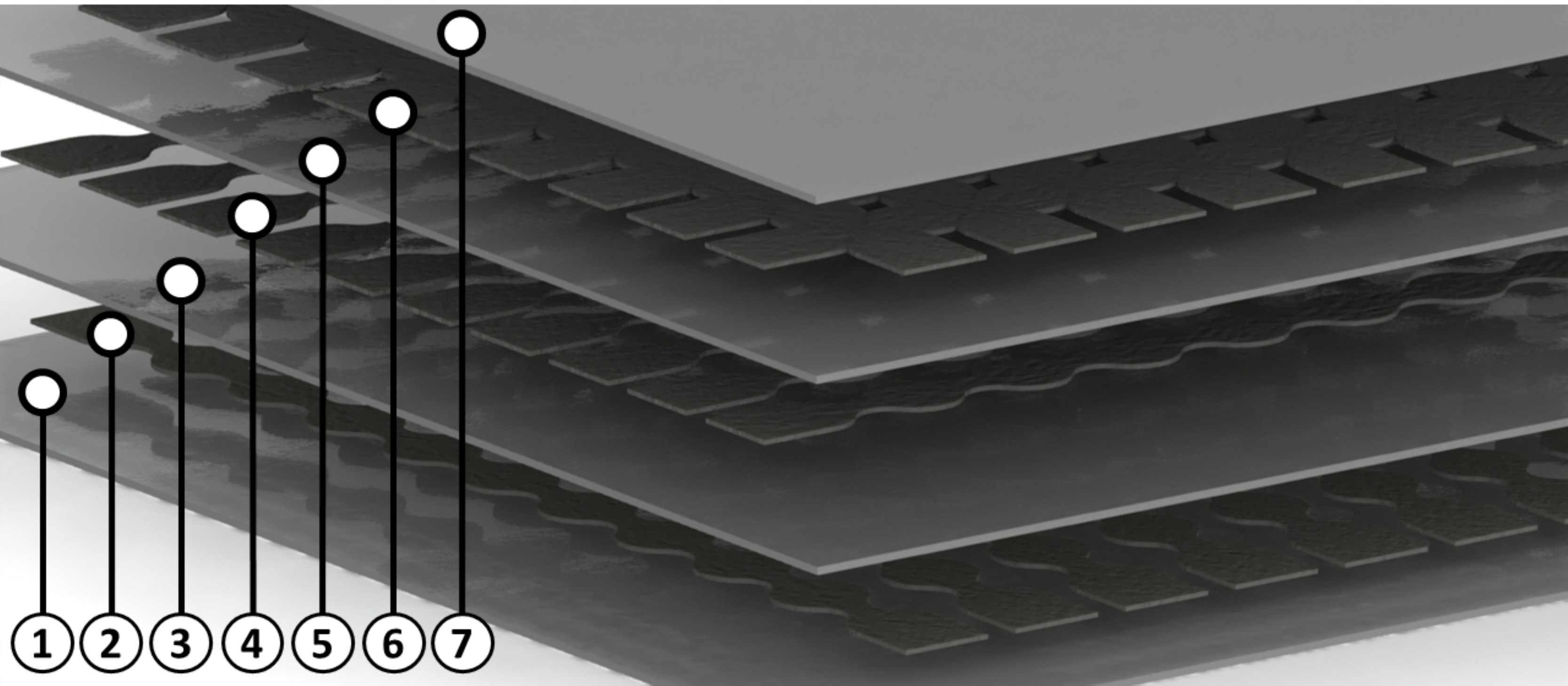}
    \caption{Exploded view of the capacitive E-skin. Labels 1, 3, 5 and 7 show the Eco-Flex layers. Label 2 shows the first layer of terminals i-e X terminals, Label 4 shows the second layer of terminals i-e Y terminals and 6 shows the third terminal layer or ground layer which is also a combination of X and Y terminals.}
    \label{fig:exploded_final_view}
\end{figure}   

\subsection{Design and Fabrication}

Our capacitive e-skin is made of three layers of terminals: a first layer of ten x-terminals, a second layer of ten y-terminals and a third layer that is connected to ground. Terminals are designed with an interlocking pattern as this has been shown to improve sensitivity \cite{Lee2014}. The exploded view of the skin with its attached terminals is shown in \Cref{fig:exploded_final_view}.

We used off-the-shelf carbon grease to make the terminals by brushing them in a stencil. These terminals are enveloped in between four Ecoflex-00-50 layers of 0.5 mm thickness. The connection between the carbon grease and external circuitry is made using  0.07 mm thick copper tape terminals. The detailed fabrication procedure is discussed in our previous work \cite{Dawood2020a}. The complete fabricated skin is shown in \Cref{fig:title_image}.

\subsection{Experimental Setup}
The experimental setup consisted of two 3D-printed clamps with an extendable bed to clamp and support the capacitive e-skin. The extendable bed underneath the e-skin provides support, doesn't let the skin stretch in downward direction hence, helps in having more local deformation contrary to our previous work \cite{Dawood2020a}.

\begin{figure}[H]
    \includegraphics[width=3.35in]{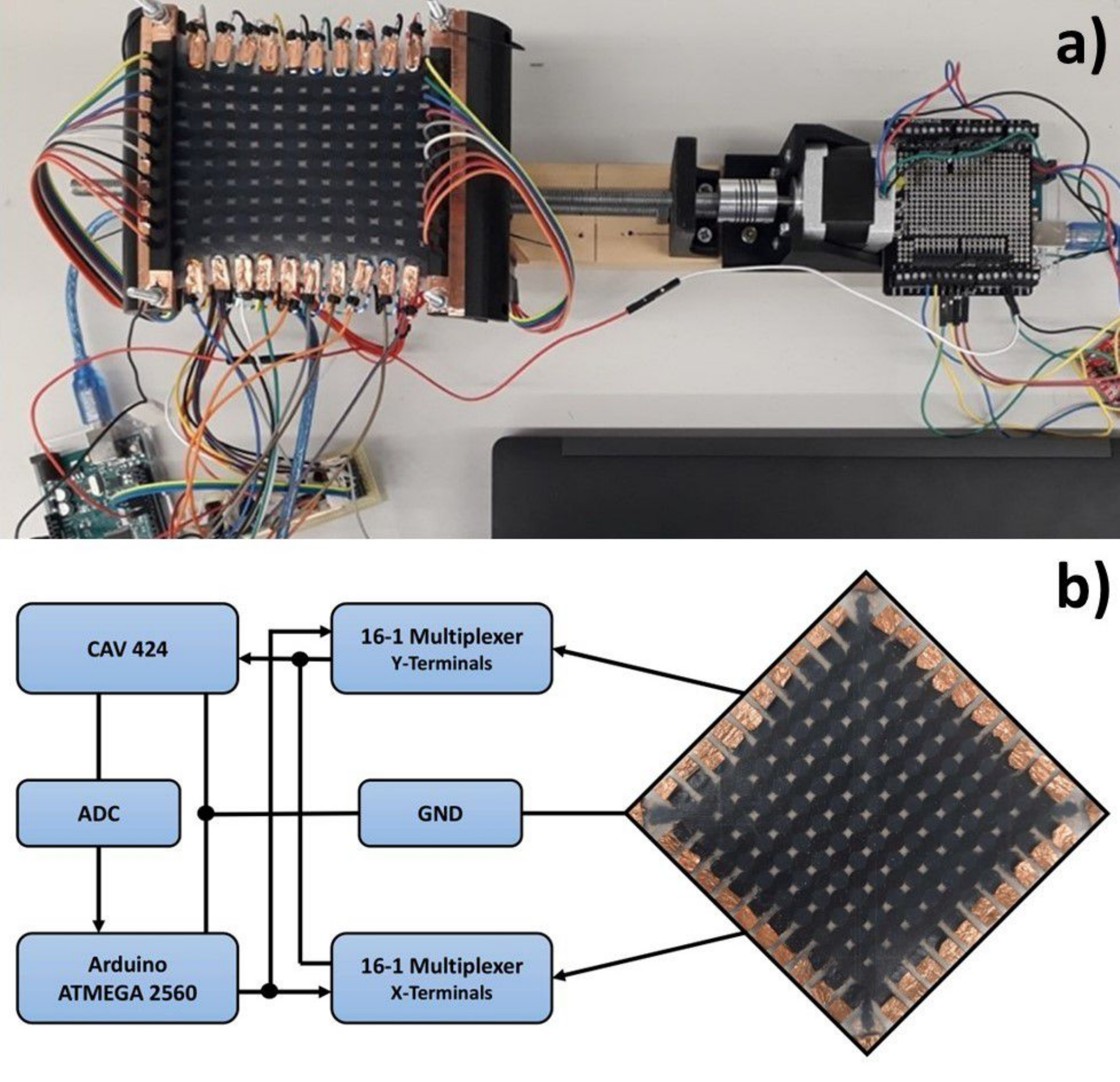}
    \caption{Experimental setup and flow-chart. a) shows the clamped capacitive skin with the stretching mechanism. b) shows the data acquisition flow chart showing the terminal connections to CAV-424 through Multiplexers, and then to the ADC of Arduino.}
    \label{fig:setup_flowchart}
\end{figure}

The 3D printed clamps also provided connections between the copper tape terminals and the wires to the circuitry. One of the clamps was held stationary on a wooden base. The other clamp was movable through an M8 lead-screw and a SUNCOR stepper motor, driven by an Arduino microcontroller. The extendable bed was designed such that a rigid support was present under the e-skin in all stretch states. This enables the e-skin to be deformed locally when subjected to indentation. The setup is shown in  \Cref{fig:setup_flowchart}a).

All twenty terminals of the e-skin were connected to an integrated circuit (IC) CAV424 via a couple of 16-1 multiplexers. CAV424 measures capacitance and converts that to a voltage that was measured using the Analog to Digital Converter (ADC) of an Arduino. Apart from reading the ADC input, the Arduino was used to switch the multiplexer, connecting each terminal to the CAV424 sequentially. The experimental setup and the flowchart of the electronics are shown in Figure \ref{fig:setup_flowchart}b.

\section{Experiments and Methods}
In this section, we discuss the experimental protocol for acquiring and processing data and how that data was used to estimate stretch as well as the force and location of local indentations. All the estimating methods were performed using the models available in Python library \texttt{scikit-learn}~\cite{sklearn_api}.

\subsection{Data Acquisition}

A 3D-printed flat indenter with a square cross-section of 7mm x 7mm was used to locally indent the sensor skin. The weight of the indenter assembly was 132 g; during experiments, the indenting weight was increased in increments of 200 g. Hence, the four force levels used in our experiments are 0N, 1.2936N, 3.2536N and 5.2136N. The e-skin had 10 $x$ and 10 $y$ terminals, creating a sensor matrix of 100 overlapping intersection, i.e., 100 nodes. The state in which no force was applied was labeled as node 0 hence, a total of 101 nodes. For each force level and for 101 nodes, 20 samples were collected resulting in 4 x 101 x 20 $=$ 8080 samples. In a further experiment, the capacitive skin with an initial length of 101mm was stretched in increments of 8mm resulting in 3 instances of the stretched sensor skin at $\lambda$ = 1, 1.07921, 1.15842. Hence, the total number of samples collected for single force applications is 3 x 8080 = 24240.

For 2-point force applications, two indenters of equal dimensions were used. It has previously been observed that an indenter, when applied to a specific node, affects two neighbouring nodes in all directions \cite{Dawood2020a}. It also means that if 2 indenters are used, there has to be at least 4 nodes between them to avoid interference. Hence, the $x$ terminal values (1, 6, 10) and $y$ terminal values (1, 6, 10) were selected resulting in 9 nodes being used, when applying 2 forces simultaneously.

In our previous work \cite{Dawood2020a}, we use similarly structured data to develop a heuristic algorithm that decouples stretch, indentation and localization measurements. 
In this work, we use the capacitive readings to develop a machine learning pipeline capable of estimating the stretch of the skin the force applied on the sensor and the coordinates of the node.

\subsection{Data Processing}

Each sample in our single force data consisted of 20 features and 4 labels. 
Twenty features correspond to 20 capacitance values for 10 $x$ and 10 $y$ terminals. Remaining four labels were force applied, $x$ and $y$ terminal numbers for node localisation and stretch. 

For 2-force application data,  each sample also consists of 26 values. 
The first 20 values have capacitance information while the next six values are the forces applied and the terminal numbers, i.e., force 1, $x$1, $y$1 along with the force 2, $x$2 and $y$2.

\subsection{Single Force Application}

\begin{figure}[t]
\centering
    \includegraphics[width=.8\columnwidth]
    {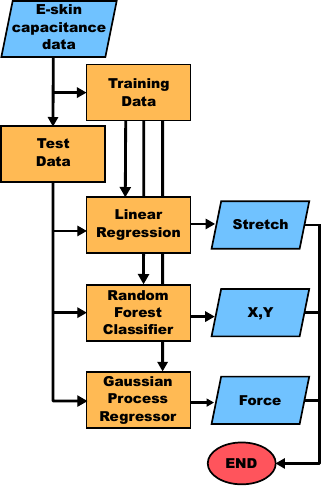}
    \caption{
    The flowchart of the proposed machine learning pipeline. The flow chart shows how the data flows through the three models: A Linear Regressor is used for stretch estimation, a Random Forest Classifier is used of localization of x and y coordinates of the node and a Gaussian Process is used for force estimation.}
    \label{fig:flow_chart_ML}
\end{figure}

We developed a machine learning algorithm for stretch estimation, force regression and localisation. The model is explained below and is shown in \Cref{fig:flow_chart_ML}.

\subsubsection{Stretch Estimation}
\label{subsec:exp_stretch}

In our previous work \cite{Dawood2019}, we have shown that the relation between the change in capacitance and stretch is linear therefore we learn the model for stretch estimation with a linear regressor. The capacitance of the row $x$ and column $y$ terminals serve as input while the stretch of the sensor is the output.
To validate this method we perform a 10-fold cross-validation stratified on the node currently pressed on the sensor to guarantee the balance of the samples for each node in the training. 

\subsubsection{Force Estimation}
The relation between change in capacitance and force is non-linear \cite{Dawood2020a}. Therefore, to estimate accurately the force of contact we use a Gaussian Process Regressor with an RBF kernel.
Gaussian Process are noise-robust and sample efficient while being able to capture non-linear relations in the data.  
The model is tested with 10-fold cross-validation on the collected dataset.

\subsubsection{Force Localisation}
\label{subsec:exp_localisation}
We localise the pressure on the sensor using a supervised classification approach. We feed the capacitance data to two classifiers: (i) a one-column classifier and (ii) a one-row classifier. Each class is associated with the corresponding row/column terminal of the sensor. The classifiers used, are Random Forest Classifiers. 
Random Forest is a reliable classification algorithm, capable of obtaining high levels of accuracy with a limited computational effort. 
The localisation algorithm assumes that there is pressure on the sensor. To operate the proposed system in a real scenario, it is essential to detect when this happens. To do so, we perform one additional classification step. 
In this case, we train an RBF Support Vector Machine (SVM) algorithm using samples with no contact as negative and all the other samples as positive. SVMs are fast and can be adjusted to compensate for imbalances in the data classes. These are important considering that this step will have to be applied repetitively before localisation. We test these components with a 10-fold cross-validation, similar to that in \Cref{subsec:exp_stretch}.%
\subsection{Two-Force application}

As mentioned earlier, nine nodes on the e-skin were selected and two forces were applied to these nodes alternately. The same algorithms were used as in the single-force instance (i.e., Gaussian Process regression for force estimation and Random Forest classifiers for localisation of the applied forces). The results are discussed in the next section.


\begin{figure}[t]
    \centering
    \includegraphics[width=\columnwidth, height=5.7cm]{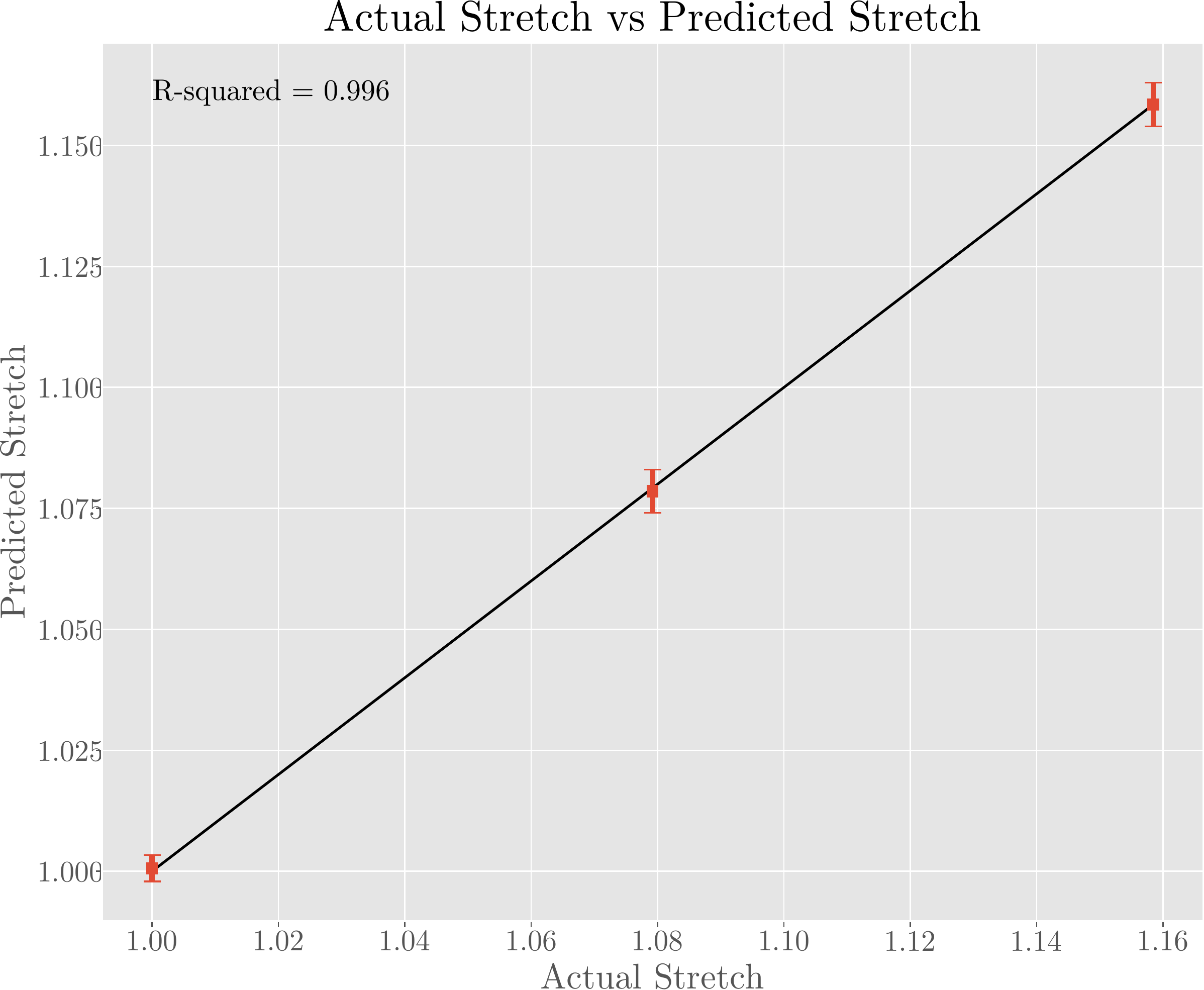}
    \caption{This Figures shows the results of our stretch estimation 
    The black line indicates the desired value of the prediction. The red markers indicate mean and standard deviation of the predictions at three different stretch levels.
    }
    \label{fig:stretch_estimation}
\end{figure}   

\begin{figure}[t]
\centering
\includegraphics[width=\columnwidth]{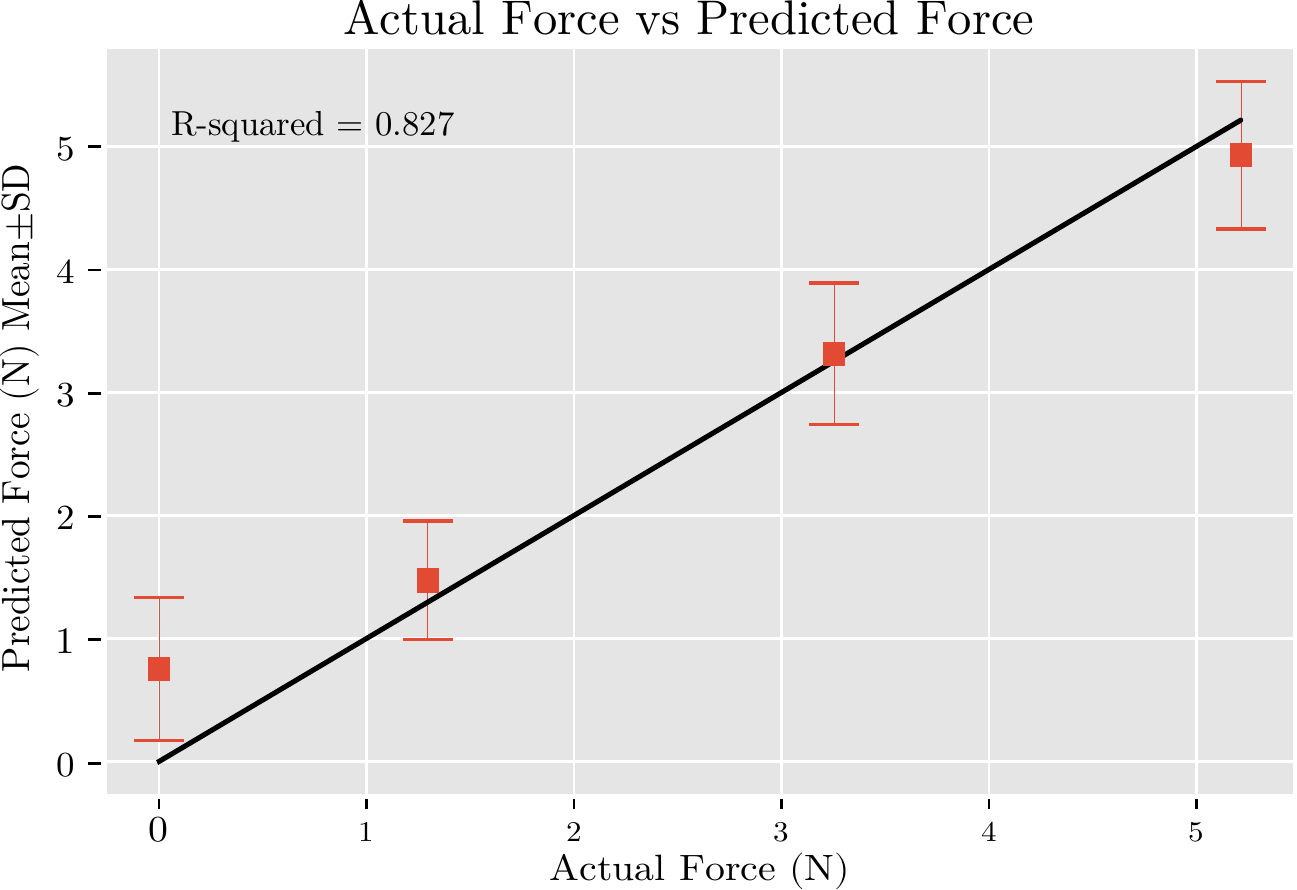}
\caption{This figure shows the results of our force estimation on the skin sensor. The black line indicates the desired value of the prediction. The red markers indicate mean and standard deviation of the predictions at three different stretch levels.}
\label{fig:force_estimation}
\end{figure}   

\begin{figure*}[t]
    \centering
    \begin{subfigure}[b]{.40\textwidth}
        \centering
        \includegraphics[width=0.99\columnwidth]{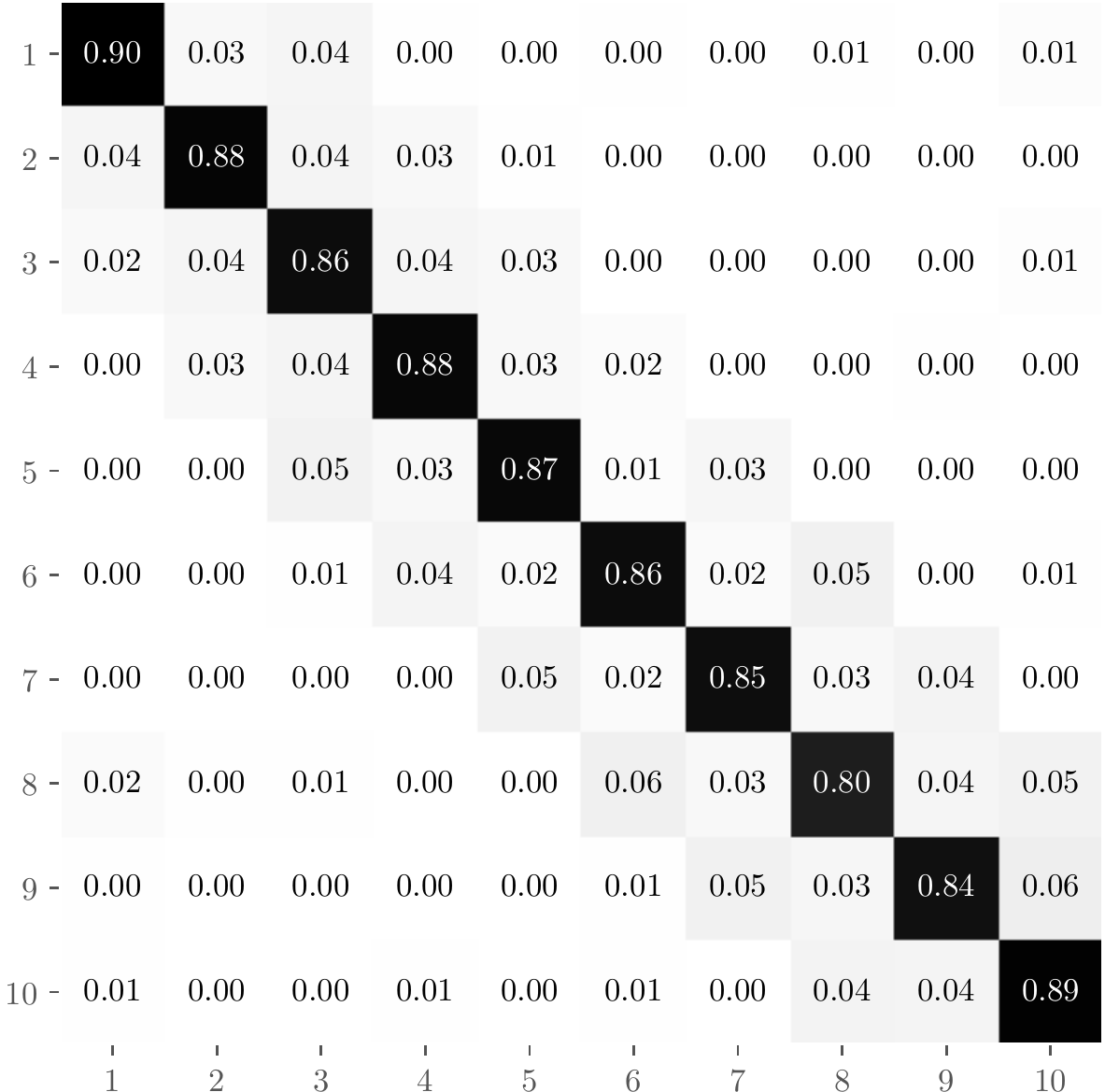}
        \caption{Confusion matrix for the localisation of the pressure on the sensor (rows).}
        \label{fig:F2_localization}
    \end{subfigure}
    \hspace{1.2cm}
    \begin{subfigure}[b]{.40\textwidth}
        \centering
        \includegraphics[width=0.99\columnwidth]{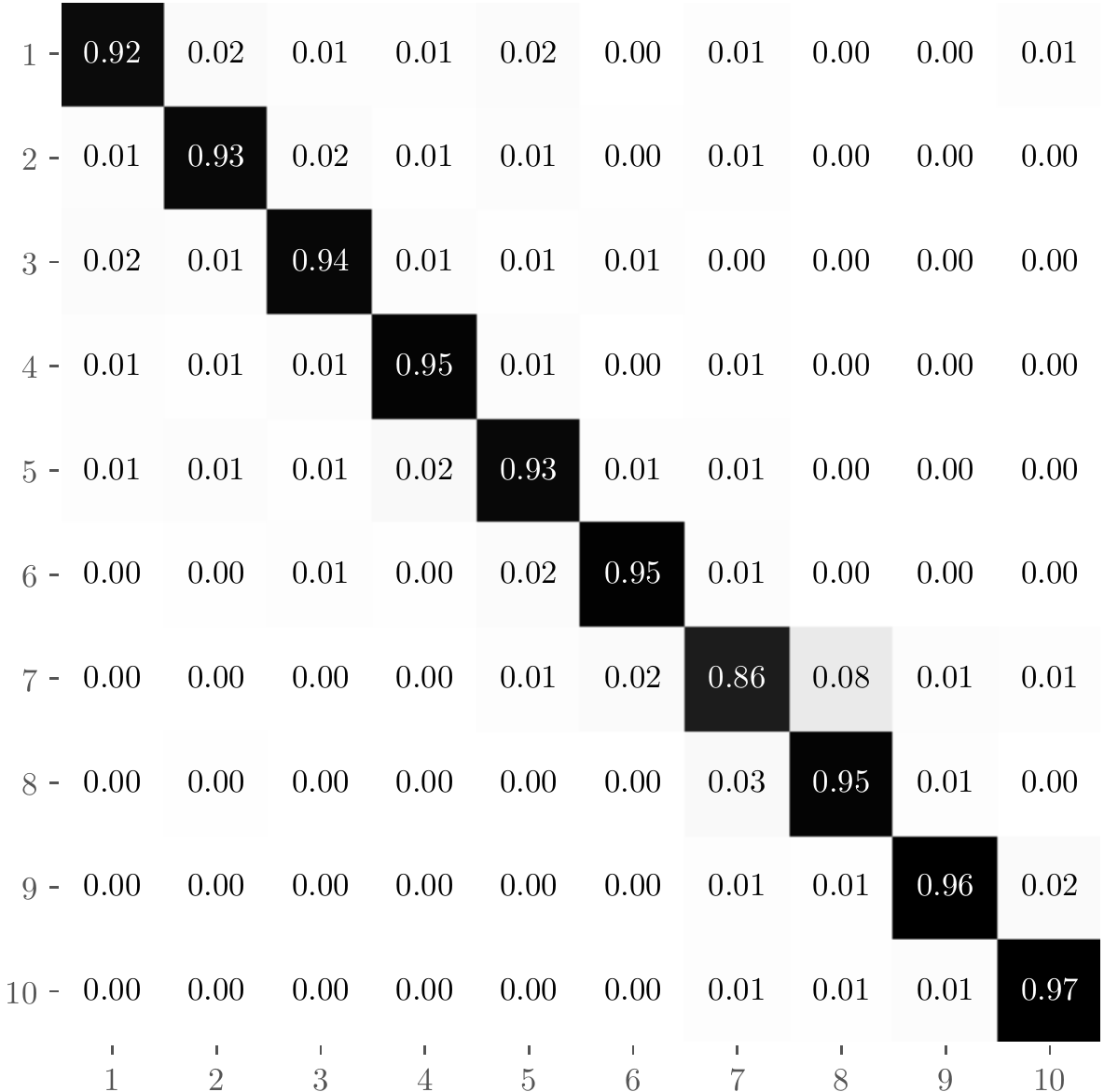}
        \caption{Confusion matrix for the localisation of the pressure on the sensor (columns).}
        \label{fig:F1_localization}
    \end{subfigure}
    \hfill
    \caption{Confusion matrices regarding the localisation accuracy}
    \label{fig:single-touch_localisation}
\end{figure*}

\section{Results and Discussion}
\subsection{Single Force Application}
Again, as already outlined, for single force application the skin was stretched to three different stretch levels, and four forces were applied to each node at each stretch level.

It has been documented in the literature that the stretch and the change in capacitance are linearly related \cite{Dawood2019}. In \Cref{fig:stretch_estimation}, we show the comparison between actual and predicted stretch obtained with linear regression on the e-skin data. The stretch is estimated with $MSE\,=\,2\times10^{-05}$ and $R^2\,=\,0.996$.

As force and the change in capacitance are non-linearly related to each other, Gaussian Process Regression was used for force estimation with an $R^2$ value of $0.827$. The results are shown in \Cref{fig:force_estimation}. The Figure shows that the model predicts the higher forces precisely, though in the case of lower forces the predicted values are not accurate.

The pressure detection algorithm is able to detect contacts with an accuracy of $96\%$.
The localisation results can be observed in \Cref{fig:single-touch_localisation}. 
They indicate that the force contact can be detected effectively by classifying capacitance values on rows/columns terminals.

\subsection{Two-Force application}
\begin{figure}[t]
    \centering
    \begin{subfigure}[b]{0.9\columnwidth}
    \centering
        \includegraphics[width=\columnwidth]{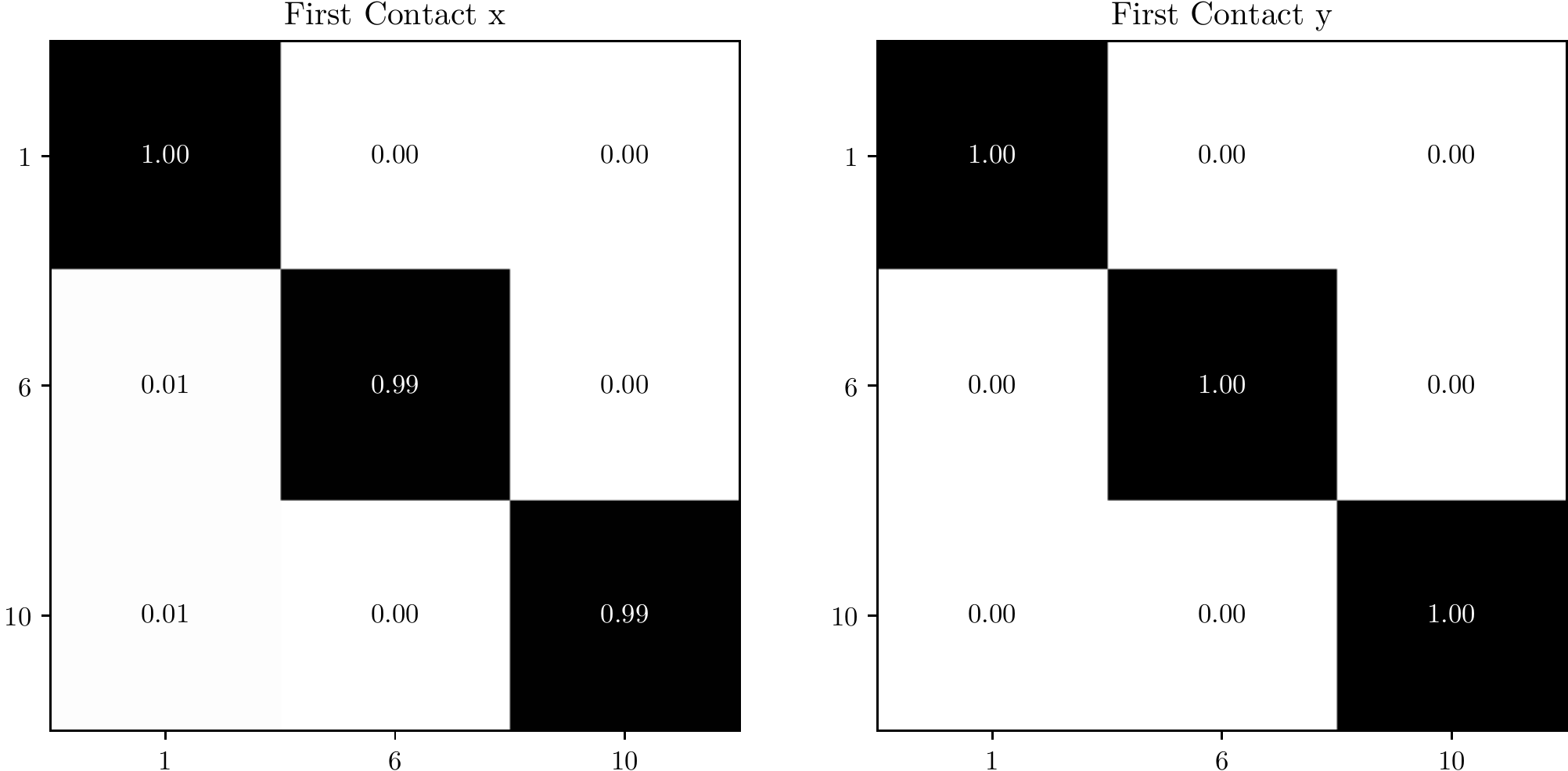}
    \end{subfigure}
    \par\bigskip
    \begin{subfigure}[b]{0.9\columnwidth}
    \centering
        \includegraphics[width=\columnwidth]{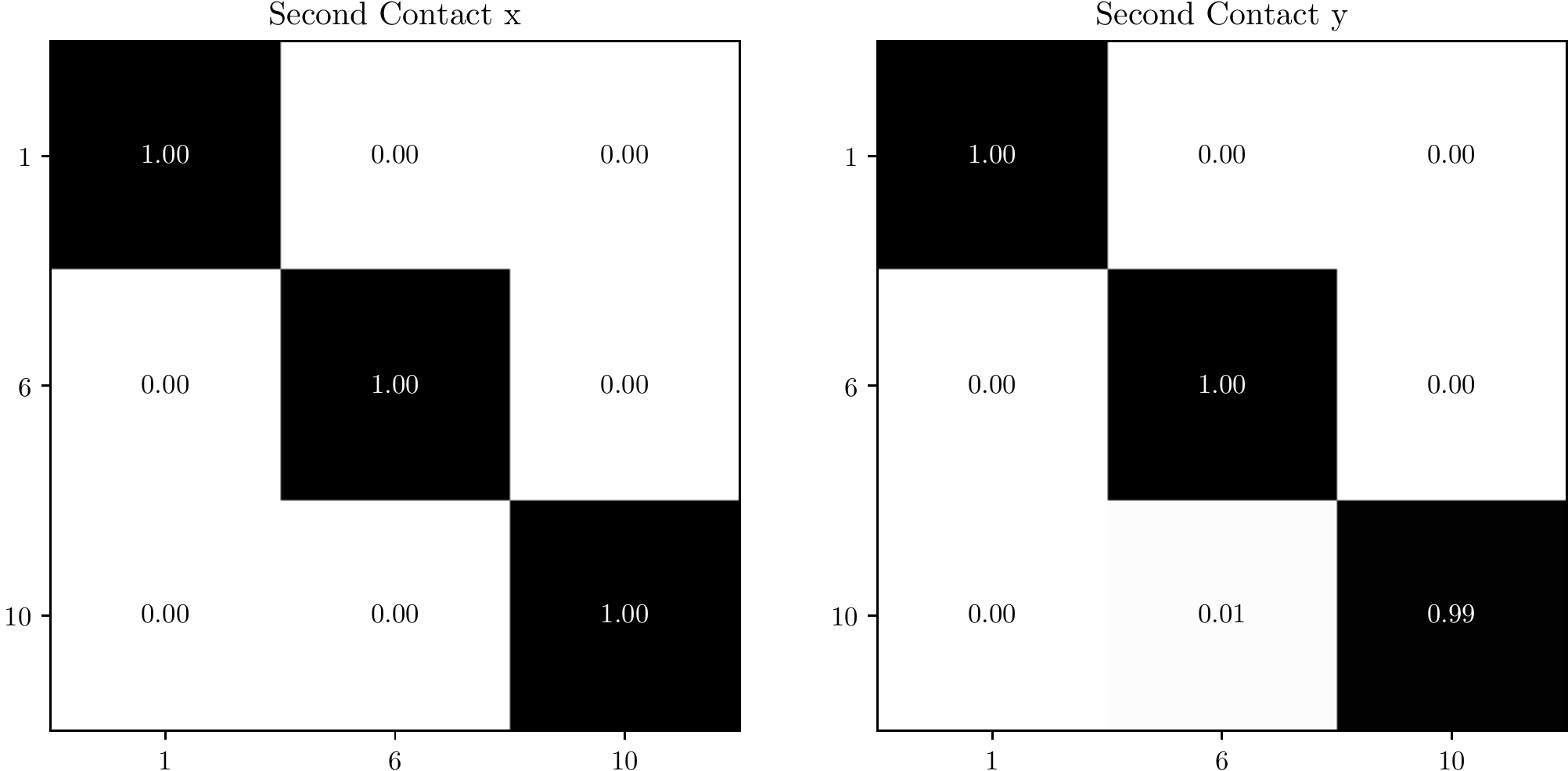}
    \end{subfigure}
    \caption{Results of multi-contact localisation experiment. On the first row, we can see the confusion in detecting the x and y of the first indenter, while on the second row we can see the confusion in detecting x and y of second indenter.}
    \label{fig:mt_localisation}
\end{figure}

At zero-stretch state, two concurrent forces are applied to the e-skin in different positions. The confusion matrices for single force application in Figure \ref{fig:single-touch_localisation} also show the affected neighboring nodes hence, proving our deduction from the previous work to be right. In \Cref{fig:mt_localisation}, we show the result of our localisation approach in the case of double contact. It can be observed that the localisation accuracy of the e-skin when two forces are applied simultaneously is high. 

\begin{figure*}[t]
    \centering
    \begin{subfigure}[t]{\columnwidth}
    \centering
    \includegraphics[width=\textwidth]{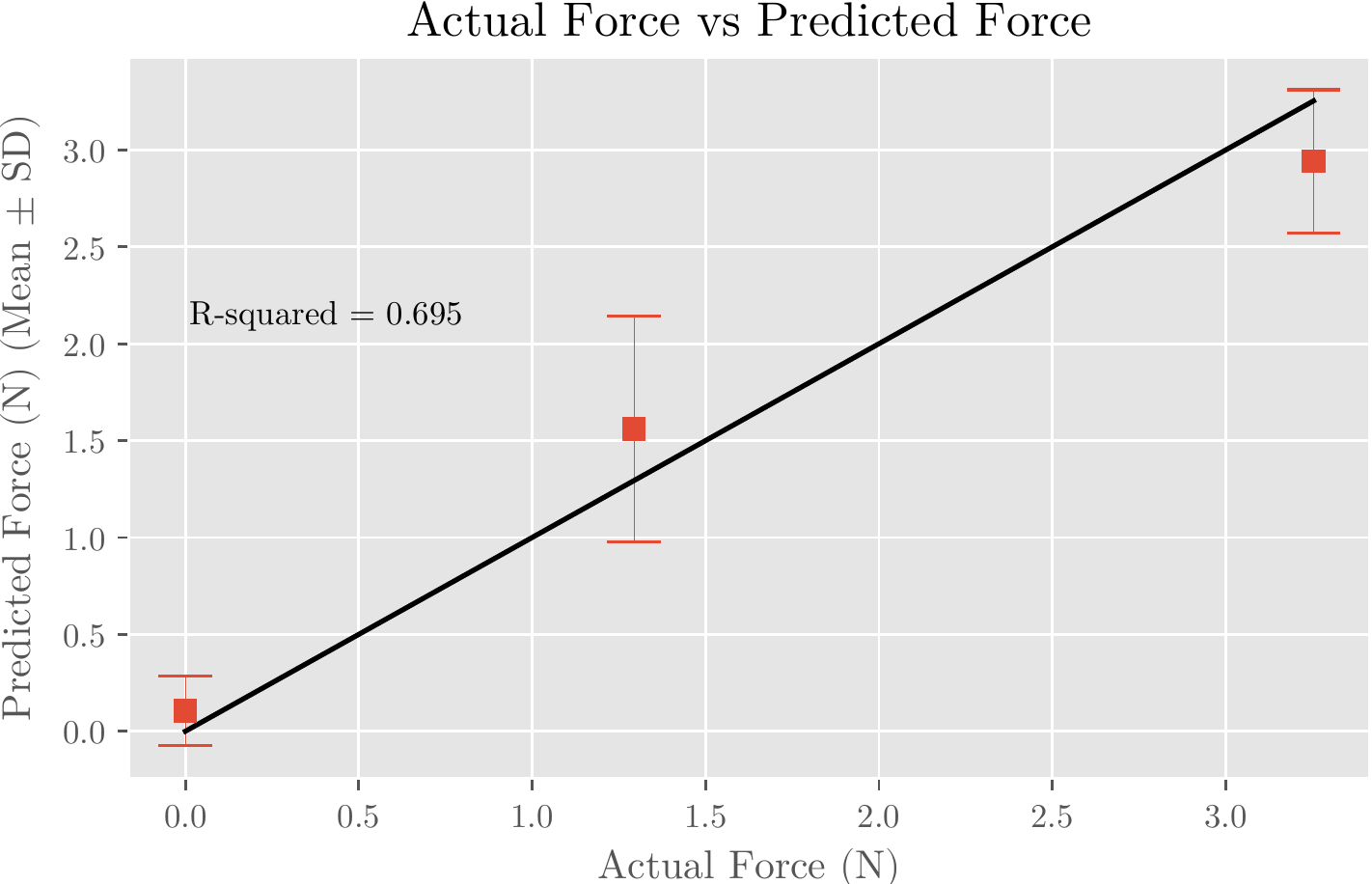}
    \caption{First Contact}
    \label{fig:mt_fRegression_main}
    \end{subfigure}
    \hfill
    \begin{subfigure}[t]{\columnwidth}
    \centering
    \includegraphics[width=\textwidth]{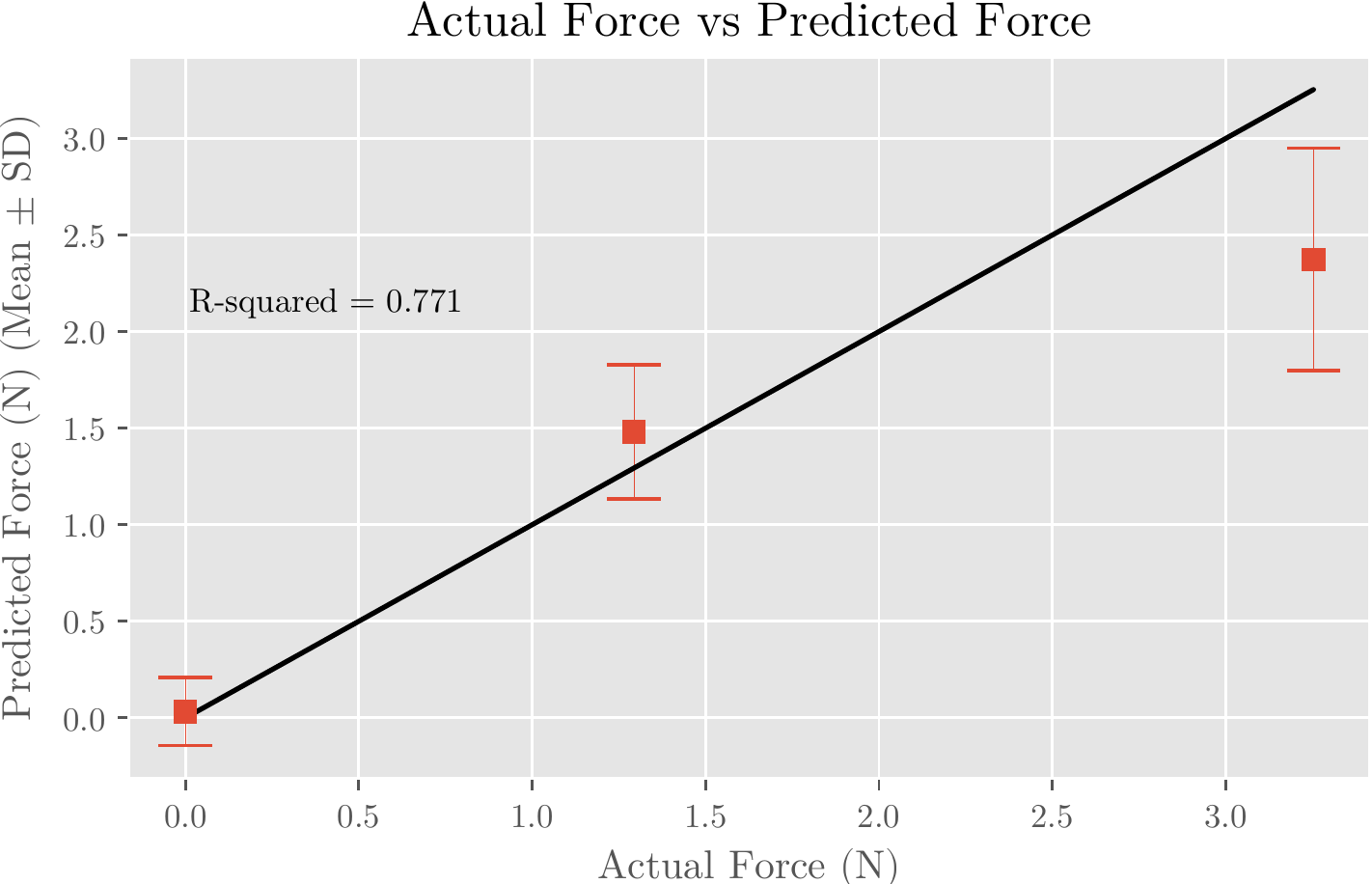}
    \caption{Second Contact}
    \label{fig:mt_fRegression_second}
    \end{subfigure}

    \caption{Results of the multitouch experiment for force regression. \textit{(a)} shows the mean and standard deviation for the force prediction of the main contact and \textit{(b)} shows the same for the second contact.}
    \label{fig:mt_fRegression}
\end{figure*}

The force estimation for the primary force and secondary force is shown in \Cref{fig:mt_fRegression}. The $R^2$ values for these forces are $0.695$ and $0.771$ respectively. 

Our experimental protocol for the two-force application was designed to have at least 4 nodes in between the two indenters. This was to minimize the effect of one indenter on the neighbouring nodes. Our results show that this experimental protocol worked well for localisation. However, for force regression, each force applied changes the capacitance of one row and one column. If the second force is applied on a node on one of those rows or columns, the capacitance is affected twice, and the  prediction when forces are applied at the same time would not be precise. This could be countered by increasing the number of samples for two force applications and will be looked into in our future work.





\section{CONCLUSIONS}

For this paper, we worked on a data-oriented approach to force and stretch estimation and localisation on a soft capacitive e-skin. Furthermore, we extend our experiments to double force application on specific nodes of the e-skin and study the effect of two forces being applied at the same time. 
We modified our experimental setup to allow our previously made capacitive e-skin to measure force, instead of an indentation. We collected the data for force application at 100 taxels and at different stretch levels. A machine learning model was used that comprised a linear regressor for stretch estimation, a Gaussian Process regressor for force estimation and an SVM along with a Random Forest Classifier for contact detection and force localisation. R$^2$ values for stretch and force estimation with our model are 0.996 and 0.827 respectively and the contact detection accuracy is 96\%. The two-force application in our experimental protocol produced good results for localisation, though  less impressive results for force regression. Future work would therefore need to focus on the improvement of this force regression model and the integration of our e-skin with a soft robot for curvature estimation and detection of external stimuli.










\bibliographystyle{IEEEtran}
\bibliography{references}

\end{document}